\documentclass{article}
\usepackage{spconf,amsmath,graphicx}
\usepackage{color}
\usepackage{boxhandler}
\usepackage{amssymb}
\usepackage[colorlinks=true, allcolors=blue]{hyperref}
\usepackage{multirow}
\usepackage[table]{xcolor}
\usepackage{array}
\usepackage{float}

\title{Skeletal movement to color map: a novel representation for 3D action recognition with inception residual networks}
\name{Huy-Hieu~Pham$^{\dag,\dag\dag,\star}$, Louahdi~Khoudour$^\dag$, Alain~Crouzil$^{\dag\dag}$, Pablo~Zegers$^{\dag\dag\dag}$, Sergio~A. Velastin$^{\dag\dag\dag\dag,\dag\dag\dag\dag\dag}$ \thanks{$^\star$This work was done when H. Pham was a visiting scholar at Carlos III University of Madrid. The authors would like to thank the financial support provided by the Cerema Research Center and Institut de Recherche en Informatique de Toulouse, Universit\'e de Toulouse, UPS, France for this research.}}
\address{$^\dag$ Cerema, Equipe-projet STI, 1 Avenue du Colonel Roche, 31400, Toulouse, France \\ $^{\dag\dag}$ Institut de Recherche en Informatique de Toulouse, Universit\'e de Toulouse, UPS, 31062 Toulouse, France\\ $^{\dag\dag\dag}$ Aparnix, La Gioconda 4355, 10B, Las Condes, Santiago, Chile \\ $^{\dag\dag\dag\dag}$ Applied Artificial Intelligence Research Group, Carlos III University of Madrid, 28270 Madrid, Spain \\ $^{\dag\dag\dag\dag\dag}$ School of Electronic Engineering and Computer Science, Queen Mary University of London, UK}
\begin{document}
\maketitle
\begin{abstract}
We propose a novel skeleton-based representation for 3D action recognition in videos using Deep Convolutional Neural Networks (D-CNNs). Two key issues have been addressed: First, how to construct a robust representation that easily captures the spatial-temporal evolutions of motions from skeleton sequences. Second, how to design D-CNNs capable of learning discriminative features from the new representation in a effective manner. To address these tasks, a skeleton-based representation, namely, SPMF (\textit{\textbf{S}keleton \textbf{P}ose-\textbf{M}otion \textbf{F}eature}) is proposed. The SPMFs are built from two of the most important properties of a human action: postures and their motions. Therefore, they are able to effectively represent complex actions. For learning and recognition tasks, we design and optimize new D-CNNs based on the idea of Inception Residual networks to predict actions from SPMFs. Our method is evaluated on two challenging datasets including MSR Action3D and NTU-RGB+D. Experimental results indicated that the proposed method surpasses state-of-the-art methods whilst requiring less computation. 
\end{abstract}
\section{Introduction}
\label{sec:1}
\begin{figure}[htb]
\begin{minipage}[b]{1.0\linewidth}
  \centering
  \centerline{\includegraphics[width=9cm,height=2.25cm]{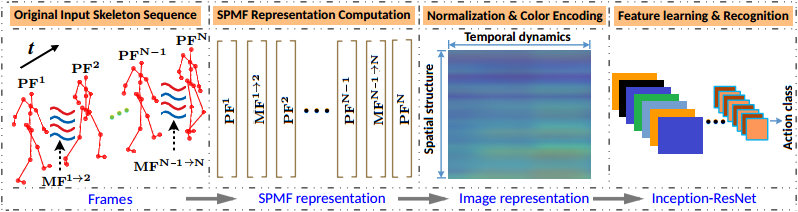}}
\end{minipage}
\caption{Schematic overview of our method. Each skeleton sequence is encoded into a color image via a skeleton-based representation called \textbf{SPMF}. Each \textbf{SPMF} is built from pose vectors (\textbf{PF}s) and motion vectors (\textbf{MF}s). They are then fed to a D-CNN, which is designed  based on the combining of Residual learning \cite{He2016DeepRL} and Inception architecture \cite{Szegedy2016RethinkingTI} for learning discriminative features from color-coded SPMFs and performing action classification.}
\label{fig:1}
\end{figure}
Recognizing correctly actions in untrimmed videos is an important but challenging problem in computer vision. The core difficulties  are occlusions, viewpoint, lighting condition and so on \cite{Poppe2010ASO}. Recently, cost-effective and easy-to-use depth cameras, \textit{e.g.} Microsoft Kinect $^{\text{TM}}$sensor or Intel RealSense, have integrated the real-time skeleton tracking algorithms \cite{Shotton2011RealTimeHP}, providing 3D structural information of human motion. This data source is robust to illumination changes, also invariant to camera viewpoints. Thus, exploiting skeletal data for 3D action recognition becomes a very effective research direction. Previous works on Skeleton-based Action Recognition (SAR) can be divided into two main groups: SAR using hand-crafted features and SAR with deep learning networks. The first group \cite{Lv2006RecognitionAS,Wang2012MiningAE,Vemulapalli2014HumanAR} uses hand-crafted local features and probabilistic graphical models such as Hidden Markov Model (HMM) \cite{Lv2006RecognitionAS}, Conditional Random Field (CRF) \cite{Han2010DiscriminativeHA}, or Fourier Temporal Pyramid (FTP) \cite{Vemulapalli2014HumanAR} to classify actions. Almost all of these approaches are data-dependent and require a lot of feature engineering. The second group \cite{Liu2016SpatioTemporalLW,Du2015HierarchicalRN,Shahroudy2016NTURA} considers SAR as a time-series problem and proposes the use of Recurrent Neural Networks with Long-Short Term Memory units (RNN-LSTMs) \cite{Hochreiter1997LongSM} to model the contextual information of the skeletons. Although RNN-LSTMs are able to model the long-term temporal of motion and have advanced the state-of-the-art, this approach just considers skeleton sequences as a kind of the low-level feature, by feeding raw skeletons directly into the network input. The huge number of input features makes RNNs become very complex and may easily lead to overfitting. Moreover, many RNN-LSTMs act as a classifier and can not extract high-level features \cite{Sainath2015ConvolutionalLS} for recognition tasks.\\
\hspace*{0.3cm} We believe that an effective representation of motion is the key factor influencing the performance of a SAR model. Different from previous studies, this paper introduces a novel representation based on skeletal data for 3D action recognition with D-CNNs. To best represent the characteristics of 3D actions, we exploit body poses (\textit{\textbf{P}ose \textbf{F}eatures} - \textbf{PF}s) and their motions (\textit{\textbf{M}otion \textbf{F}eatures} - \textbf{MF}s) for building a new representation called \textbf{SPMF} (\textit{\textbf{S}keleton \textbf{P}ose-\textbf{M}otion \textbf{F}eature}). Each SPMF contains important characteristics related to the spatial structure and temporal dynamics of skeletons. A new deep framework based on the Inception Residual networks (Inception-ResNets \cite{Szegedy2017Inceptionv4IA}) is then proposed for learning and classifying actions, as shown in Figure~\ref{fig:1}. Generally speaking, four hypotheses motivate our exploration of using D-CNNs for SAR are: \textbf{(1)} human actions can be correctly represented via the movement of skeletons; \textbf{(2)} the spatial-temporal dynamics of skeletons can be transformed into a 2D image structure -- a representation that can be effectively learned by learning methods as D-CNNs; \textbf{(3)} compared to RGB and depth modalities, skeletons are high-level information while with much less complexity than RGB-D sequences, making action learning model much simpler and faster; (\textbf{4}) a well-designed and deeper CNN can improve learning accuracy. Our experimental results on two benchmark datasets confirmed these statements.\\
\hspace*{0.3cm} The main contributions of this paper can be summarized as follows: \textbf{First}, we investigate and propose a novel skeleton-based representation, namely SPMF, for 3D action recognition. \textbf{Second}, we design and optimize new D-CNNs based on Inception-ResNet for learning motion features from SPMF in an end-to-end learning framework. To the best of our knowledge, this is the first time a very deep network as Inception-ResNet is exploited for SAR. \textbf{Last}, the proposed method set a new state-of-the-art on the MSR Action3D and the NTU-RGB+D datasets. In the next section, we introduce the SPMF representation and the proposed deep learning networks. Experiments and their results are provided in Section~\ref{sect:4}. Section~\ref{sect:5} concludes the paper and discusses the future work.
\section{Proposed Method}
\label{sect:2}
\subsection{SPMF: From skeleton movement to color map}
Two key elements to determine an action are static postures and their motions. We propose SPMF, a novel representation based on these features that are extracted from skeletons. Note that, combining too many geometric features will lead to lower performance than using only a single feature or several main features \cite{Zhang2017OnGF}. In our study, each SPMF is built from pose and motion vectors, as described below:
\subsubsection{Pose Feature (PF)}
Given a skeleton sequence $\mathcal{S}$ with $N$ frames, denoting as $\mathcal{S} = \{\textbf{F}^t\}$ with $t \in [1,N]$. Let $\textbf{p}_j^t$ and $\textbf{p}_k^t$ be the 3D coordinates of the $j$-th and $k$-th joints in each skeleton frame $\textbf{F}^t$ ($j \neq k$). The \textbf{J}oint-\textbf{J}oint \textbf{E}uclidean distance $\textbf{JJD}_{jk}^t$ between $\textbf{p}_j^t$ and $\textbf{p}_k^t$ at timestamp $t$ is given by:
\begin{equation}
\textbf{JJD}_{jk}^t = ||\textbf{p}_j^t - \textbf{p}_k^t ||_2, j \neq k, t \in [1,N]
\label{eqn:1}
\end{equation}
\hspace*{0.3cm} The joint distances obtained by Eq. (\ref{eqn:1}) for the whole dataset range from $\textbf{D}_{\text{min}} = \textbf{0}$ to $\textbf{D}_{\text{max}} = \text{max}\{\textbf{JJD}_{jk}^t\}$ and noted by $\mathcal{D}_{\text{\textbf{original}}}$. The 3D action features can be learned directly from $\mathcal{D}_{\text{\textbf{original}}}$ by D-CNNs. However, as $\mathcal{D}_{\text{\textbf{original}}}$ is high-dimensional parametric input space, it may lead D-CNNs to time-consuming and overfitting. Therefore, we normalize $\mathcal{D}_{\text{\textbf{original}}}$ to the range $[\textbf{0},\textbf{1}]$, denote as $\mathcal{D}_{[\textbf{0},\textbf{1}]}$ and considering each distance value in $\mathcal{D}_{[\textbf{0},\textbf{1}]}$ as a pixel of the intensity information in a greyscale image. To reflect the change in joint distances, we encode $\mathcal{D}_{[\textbf{0},\textbf{1}]}$ into a color space by a sequential discrete color palette. The encoding process is performed by 256-color JET\footnote{ A JET color map is based on the order of colors in the spectrum of visible light, ranging from blue to red, passes through the cyan, yellow, orange.} scale. The use of a sequential discrete color palette helps us to reduce the dimensionality of input space that accelerates the convergence rate of deep networks during the representation learning later.\\
\hspace*{0.3cm} The \textbf{J}oint-\textbf{J}oint \textbf{O}rientation $\textbf{JJO}_{jk}^t$ from joint $\textbf{p}_j^t$ to $\textbf{p}_k^t$ at timestamp $t$, represented by the unit vector $\overrightarrow{\textbf{p}_j^t\textbf{p}_k^t}$ as follows:
\begin{equation}
\textbf{JJO}_{jk}^t = \text{unit}(\textbf{p}_j^t - \textbf{p}_k^t) = \frac{\textbf{p}_j^t - \textbf{p}_k^t}{\textbf{JJD}_{jk}^t} , j \neq k , t \in  [1,N]
\label{eqn:2}
\end{equation}
Each vector $\textbf{JJO}_{jk}^t$ is a 3D vector where all of its components can be normalized to the range $[\textbf{0},\textbf{255}]$. We consider three components ($x,y,z$) of $\textbf{JJO}_{jk}^t$ after normalization as the corresponding three components ($R,G,B$) of a color pixel and define a human pose at timestamp $t$ by vector $\textbf{PF}^t$ that describes the distance and orientation relationship between joints as a color matrix\footnote{ To concatenate vectors of unequal lengths, we expand shorter vectors with neighborhood values until reaching the length of the larger vectors. Before that, all duplicated distances or zero vectors are removed.}. Each $\textbf{PF}^t$ is obtained by:
\begin{equation}
 \textbf{PF}^t = \left [ \textbf{JJD}_{jk}^t \hspace{0.1cm} \textbf{JJO}_{jk}^t  \right ], {j \neq k,t \in [1, N]}
 \label{eqn:3}
\end{equation}
\subsubsection{Motion Feature (MF)}
Let $\textbf{p}_j^t$ and $\textbf{p}_k^{t+1}$ be the 3D coordinates of the $j$-th and $k$-th joints at two frames consecutively $t$ and $t + 1$. Similarly to $\textbf{JJD}_{jk}^t$ in Eq. (\ref{eqn:1}), the \textbf{J}oint-\textbf{J}oint \textbf{E}uclidean distance $\textbf{JJD}_{jk}^{t,t+1}$ between $\textbf{p}_j^t$ and $\textbf{p}_k^{t+1}$ is computed as:
\begin{equation}
\textbf{JJD}_{jk}^{t,t+1} = ||\textbf{p}_j^t - \textbf{p}_k^{t+1} ||_2 , t \in  [1,N-1]
\label{eqn:4}
\end{equation}
Also, the \textbf{J}oint-\textbf{J}oint \textbf{O}rientation $\textbf{JJO}_{jk}^{t,t+1}$ from joint $\textbf{p}_j^t$ to $\textbf{p}_k^{t+1}$, represented by the unit vector $\overrightarrow{\textbf{p}_j^t\textbf{p}_k^{t+1}}$ as follows:
\begin{equation}
\textbf{JJO}_{jk}^{t,t+1} = \text{unit}(\textbf{p}_j^t - \textbf{p}_k^{t+1}) = \frac{\textbf{p}_j^t - \textbf{p}_k^{t+1}}{\textbf{JJD}_{jk}^{t,t+1}}, t \in  [1,N-1]
\label{eqn:5}
\end{equation}
We represent a motion from frame $t$ to $t+1$ by vector $\textbf{MF}^{t\to t+1}$, given by:
\begin{equation}
\textbf{MF}^{t\to t+1} = \left [\textbf{\textbf{JJD}}_{jk}^{t,t+1} \hspace{0.1cm} \textbf{\textbf{JJO}}_{jk}^{t,t+1} \right ], {t \in [1, N-1]}
\label{eqn:6}
\end{equation}
where $\textbf{\textbf{JJD}}_{jk}^{t,t+1}$ and $\textbf{JJO}_{jk}^{t,t+1}$ are encoded to color matrices as $\textbf{\textbf{JJD}}_{jk}^{t}$ and $\textbf{JJO}_{jk}^{t}$. This process not only enhances the texture information in image representation, but also distinguishes the slower and faster motions based on their colors.   
\subsubsection{Modeling human action with PFs and MFs}
\begin{figure}[htb]
\begin{minipage}[b]{1.0\linewidth}
  \centering
  \centerline{\includegraphics[width=8.6cm,height=1.7cm]{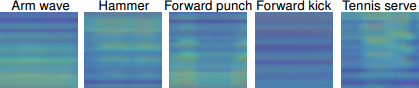}}
\end{minipage}. \caption{The SPMFs obtained from some samples of the MSR Action3D dataset. Color-changing reflects the change in distance between skeleton joints. Best viewed in color.}
\label{fig:SPMFs}
\end{figure}

Based on the PFs and MFs, we propose a novel skeleton-based representation, namely \textbf{SPMF} (\textit{\textbf{S}keleton \textbf{P}ose-\textbf{M}otion \textbf{F}eatures}). To build the SPMFs, the PFs and MFs are concatenated into a single feature vector by temporal order. Specifically, a SPMF represents a skeleton sequence $\mathcal{S}$, without dependence on the range of actions, is defined as:
\begin{equation}
\begin{split}
\textbf{SPMF}_{\mathcal{S}}=[\textbf{PF}^{1} \hspace{0.08cm}  \textbf{MF}^{1\to2} \hspace{0.08cm}  \textbf{PF}^{2} \hspace{0.08cm} ... \hspace{0.08cm} \textbf{PF}^{t} \hspace{0.08cm} \textbf{MF}^{t\to t+1} \hspace{0.08cm} \hspace{0.08cm}  \textbf{PF}^{t+1} \\
...\hspace{0.08cm} \textbf{PF}^{N-1} \hspace{0.08cm} \textbf{MF}^{N-1\to N} \hspace{0.08cm} \textbf{PF}^{N}], {t \in [2, N-2]} \\
\end{split}
\label{eqn:7}
\end{equation}
Figure~\ref{fig:SPMFs} shows some SPMFs obtained from the MSR Action3D dataset \cite{Li2010ActionRB} after resizing them to $32\times32$ pixels.
\subsection{Inception Residual learning}
%Inception architecture
D-CNNs have demonstrated state-of-the-art performance on many visual recognition tasks. In particular, the recent Inception architecture \cite{Szegedy2016RethinkingTI} significantly improved both the accuracy and computational cost through three key ideas: (\textbf{1}) reducing the input dimension; (\textbf{2}) increasing  not only the network depth, but also its width and (\textbf{3}) concatenating feature maps learned by different layers. However, very deep networks as Inception are very difficult to train due to the vanishing problem and degradation phenomenon \cite{He2015ConvolutionalNN}. To this end, ResNet \cite{He2016DeepRL} has been introduced. The key idea is to improve the flow of information and gradients through layers by using identity connections. A layer, or a sequence of layers of a traditional CNN learns to calculate a mapping function  $y = \mathcal{F}(x)$ from the input feature $x$. Meanwhile, a ResNet building block approximately calculates the function $y = \mathcal{F}(x) + id(x)$ where $id(x) = x$. This idea helps the learning process to be faster and more accurate. To learn spatio-temporal features from the SPMFs, we propose the combination of Residual learning \cite{He2016DeepRL} and Inception architecture \cite{Szegedy2016RethinkingTI} to build D-CNNs\footnote{ Please refer to the Supplementary Materials to see details of the proposed network architectures.}. Batch normalization \cite{Ioffe2015BatchNA} and Exponential Linear Units (ELUs) \cite{Clevert2015FastAA} are applied after each Convolution. Dropout \cite{Hinton2012ImprovingNN} with a rate of \textbf{0.5} is used to prevent overfitting. A Softmax layer is employed for classification task. Our networks can be trained in an end-to-end manner by the gradient decent using Adam update rule \cite{Kingma2014AdamAM}. During training, our goal is to minimize the cross-entropy loss function between the ground-truth label $\textbf{y}$ and the predicted label $\hat{\textbf{y}}$ by the network over the training samples $\mathcal{X}$, which is expressed as follows:
\begin{equation}
\mathcal{L}_{\mathcal{X}}(\textbf{y},\hat{\textbf{y}}) = - \frac{1}{M}\left (\sum_{i = 1}^{M} \left (  \sum_{j = 1}^{C} \textbf{y}_{ij} \log \hat{\textbf{y}}_{ij} \right )\right )
\end{equation}
where $M$ indicates the number of samples in training set $\mathcal{X}$ and $C$ denotes the number of action classes. 
\section{Experiments}
\label{sect:4}
\subsection{Datasets and settings}
The proposed method is evaluated on the MSR Action3D and NTU-RGB+D datasets\footnote{ The list of action classes is provided in Supplementary Materials.}. We follow the evaluation protocols as provided in the original papers. The performance is measured by average classification accuracy over all action classes.\\
\\
\textbf{MSR Action3D dataset} \cite{Li2010ActionRB}: This Kinect 1 captured dataset contains 20 actions, performed by 10 subjects. Each skeleton has \textbf{20} joints. Our experiments were conducted on 557 action sequences in which the whole dataset is divided into three subsets: Action Set 1 (\textbf{AS1}), Action Set 2 (\textbf{AS2}), and Action Set 3 (\textbf{AS3}). For each subset, a half of the data is selected for training and the rest for testing.  More details are provided in the Supplementary Material.\\
\\
\textbf{NTU-RGB+D dataset} \cite{Shahroudy2016NTURA}: This Kinect 2 captured dataset is currently the largest dataset for SAR, also very challenging due to its large intra-class and multiple viewpoints. The NTU-RGB+D provides more than 56,000 videos, collected from 40 subjects for 60 action classes. Each skeleton contains the 3D coordinates of \textbf{25} body joints. Two different evaluation criteria have been suggested. For the \textbf{Cross-Subject} evaluation, the sequences performed by 20 subjects are used for training and the rest sequences are used for testing. For  \textbf{Cross-View} evaluation, the sequences provided by cameras \textbf{2} and \textbf{3} are used for training while sequences from camera \textbf{1} are used for testing. 
\subsection{Implementation details}
Three different network configurations were implemented and evaluated in Python with Keras framework using the TensorFlow backend. During training, we use mini-batches of \textbf{256} images for all networks. The weights are initialized by the He initialization technique \cite{He2015DelvingDI}. Adam optimizer \cite{Kingma2014AdamAM} is used with default parameters, $\beta_1 = \textbf{0.9}$ and $\beta_2 = \textbf{0.999}$. The initial learning rate is set to \textbf{0.001} and is decreased by a factor of \textbf{0.5} after every \textbf{20} epochs. All networks are trained for \textbf{250} epochs from scratch. We applied some simple data augmentation techniques (\textit{i.e.}, randomly cropping, flipping and Gaussian filtering) on the MSR Action3D dataset \cite{Li2010ActionRB} due to its small size. For the NTU-RGB+D \cite{Shahroudy2016NTURA}, we do not apply any data augmentation method. 
\subsection{Results and comparison with the state-of-the-art}
Table~\ref{MSR-Action3D} reports the experimental results and comparisons with state-of-the-art methods on the MSR Action3D dataset \cite{Li2010ActionRB}. We achieved the best recognition accuracy by SPMF Inception-ResNet-222 network configuration with a total average accuracy of \textcolor{blue}{\textbf{98.56}}\%. This result outperforms many previous studies \cite{Li2010ActionRB,Chen2013RealtimeHA,Vemulapalli2014HumanAR,Du2015HierarchicalRN,Liu2016SpatioTemporalLW,Wang2016GraphBS,WengSpatioTemporalNN,Xu2015SpatioTemporalPM,DBLP:journals/corr/LiWWHL17}. For the NTU-RGB+D dataset \cite{Shahroudy2016NTURA}, we achieved an accuracy of \textcolor{blue}{\textbf{78.89}}\% on cross-subject evaluation and \textcolor{blue}{\textbf{86.15}}\% on cross-view evaluation as shown in Table~\ref{NTU-Accuracy}. These results are better than previous state-of-the-art works reported in \cite{Vemulapalli2014HumanAR,Du2015HierarchicalRN,Liu2016SpatioTemporalLW, Shahroudy2016NTURA,Hu2015JointlyLH,Rahmani_2017_ICCV}. 
\begin{table}
\centering
\caption{\label{MSR-Action3D} Accuracy rate (\%) on the MSR Action3D dataset.}
%MSR-Action3D-accuracy} Model accuracy (\%) and comparison with state-the-art methods on the MSR Action3D dataset
\begin{tabular}{lllll}
\hline
\hline
 {\footnotesize \hspace*{0.01cm} \textbf{Method (protocol of \cite{Li2010ActionRB})}} & {\footnotesize \hspace*{0.01cm}  \textbf{AS1}} & {\footnotesize \hspace*{0.01cm} \textbf{AS2}} & {\footnotesize  \hspace*{0.01cm} \textbf{AS3}} &  {\footnotesize \hspace*{0.01cm} \textbf{Aver.}}  \\
\hline
{\footnotesize \hspace*{0.01cm} Bag of 3D Points \cite{Li2010ActionRB}} & {\footnotesize \hspace*{0.01cm} 72.90} & {\footnotesize \hspace*{0.01cm} 71.90}  & {\footnotesize \hspace*{0.01cm} 71.90} &  {\footnotesize \hspace*{0.01cm} 74.70} \\
{\footnotesize \cellcolor{gray!50} Depth Motion Maps \cite{Chen2013RealtimeHA}} & {\footnotesize \cellcolor{gray!50} 96.20} & {\footnotesize \cellcolor{gray!50} 83.20}  & {\footnotesize \cellcolor{gray!50} 92.00} &  {\footnotesize \cellcolor{gray!50} 90.47} \\
{\footnotesize \hspace*{0.01cm} Lie Group Representation \cite{Vemulapalli2014HumanAR}} & {\footnotesize \hspace*{0.01cm}  95.29} & {\footnotesize \hspace*{0.01cm}  83.87}  & {\footnotesize  \hspace*{0.01cm} 98.22} &  {\footnotesize \hspace*{0.01cm} 92.46} \\
{\footnotesize \cellcolor{gray!50} Hierarchical RNN \cite{Du2015HierarchicalRN}} & {\footnotesize \cellcolor{gray!50}    \textcolor{blue}{\textbf{99.33$^\dag$}}} & {\footnotesize \cellcolor{gray!50}  94.64}  & {\footnotesize \cellcolor{gray!50}   95.50} &  {\footnotesize \cellcolor{gray!50}  94.49} \\
{\footnotesize  \hspace*{0.01cm} ST-LSTM Trust Gates \cite{Liu2016SpatioTemporalLW}} & {\footnotesize \hspace*{0.01cm} N/A} & {\footnotesize  \hspace*{0.01cm} N/A}  & {\footnotesize  \hspace*{0.01cm} N/A} &  {\footnotesize  \hspace*{0.01cm} 94.80}
 \\
{\footnotesize \cellcolor{gray!50}  Graph-Based Motion \cite{Wang2016GraphBS}} & {\footnotesize \cellcolor{gray!50}   93.60} & {\footnotesize  \cellcolor{gray!50}  95.50}  & {\footnotesize  \cellcolor{gray!50}   95.10} &  {\footnotesize \cellcolor{gray!50}   94.80}
 \\
{\footnotesize  \hspace*{0.01cm} ST-NBNN \cite{WengSpatioTemporalNN}} & {\footnotesize  \hspace*{0.01cm} 91.50} & {\footnotesize   \hspace*{0.01cm} 95.60}  & {\footnotesize   \hspace*{0.01cm} 97.30} &  {\footnotesize  \hspace*{0.01cm} 94.80} \\
{\footnotesize \cellcolor{gray!50} S-T Pyramid \cite{Xu2015SpatioTemporalPM}} & {\footnotesize \cellcolor{gray!50} 99.10} & {\footnotesize \cellcolor{gray!50} 92.90}  & {\footnotesize  \cellcolor{gray!50} 96.40} &  {\footnotesize \cellcolor{gray!50} 96.10} \\
{\footnotesize \hspace*{0.01cm} Ensemble TS-LSTM v2 \cite{DBLP:journals/corr/LiWWHL17}} & {\footnotesize  \hspace*{0.01cm} 95.24} & {\footnotesize  \hspace*{0.01cm} 96.43}  & {\footnotesize   \hspace*{0.01cm}  \textcolor{blue}{\textbf{100.0}}} &  {\footnotesize  \hspace*{0.01cm} 97.22} \\
\hline
{\footnotesize \cellcolor{gray!50} SPMF Inception-ResNet-121$^{\ddag}$} & {\footnotesize \cellcolor{gray!50} 97.06} & {\footnotesize \cellcolor{gray!50}   \textcolor{blue}{\textbf{99.00}}}  & {\footnotesize \cellcolor{gray!50} 98.09} &  {\footnotesize \cellcolor{gray!50} \textbf{98.05}} \\
{\footnotesize  \hspace*{0.01cm} SPMF Inception-ResNet-222} & {\footnotesize  \hspace*{0.01cm} 97.54} & {\footnotesize  \hspace*{0.01cm} \textbf{98.73}}  & {\footnotesize  \hspace*{0.01cm} 99.41} &  {\footnotesize  \hspace*{0.01cm}  \textcolor{blue}{\textbf{98.56}}} \\
{\footnotesize \cellcolor{gray!50} SPMF Inception-ResNet-242} & {\footnotesize \cellcolor{gray!50} 96.73} & {\footnotesize\cellcolor{gray!50} \textbf{97.35}}  & {\footnotesize \cellcolor{gray!50} 98.77} &  {\footnotesize \cellcolor{gray!50}  \textbf{97.62}} \\
\hline
\hline
\end{tabular}
$^\dag$\scriptsize Results that outperform previous works are in \textbf{bold}, best accuracies are in \textcolor{blue}{\textbf{bold-blue}}.
$^{\ddag}$\scriptsize Denote the number of building blocks Inception-ResNet-A, Inception-ResNet-B, and Inception-ResNet-C, respectively. Details are provided in Supplementary Materials.
\end{table}

\begin{table}
\centering
\caption{\label{NTU-Accuracy} Accuracy rate (\%) on the NTU-RGB+D dataset.}
\begin{tabular}{lll}
\hline 
\hline
\hspace*{0.01cm} {\footnotesize \textbf{Method (protocol of \cite{Shahroudy2016NTURA})}} & {\footnotesize \textbf{Cross-Subject}} & {\footnotesize \textbf{Cross-View}}  \\
\hline
{\footnotesize \hspace*{0.01cm} Lie Group Representation \cite{Vemulapalli2014HumanAR}} & {\footnotesize \hspace*{0.32cm} 50.10} & {\footnotesize \hspace*{0.24cm} 52.80}  \\
{\footnotesize \cellcolor{gray!50} Hierarchical RNN \cite{Du2015HierarchicalRN}} & {\footnotesize \cellcolor{gray!50}  \hspace*{0.25cm} 59.07} & {\footnotesize  \cellcolor{gray!50}  \hspace*{0.17cm} 63.97}  \\
{\footnotesize \hspace*{0.01cm} ST-LSTM Trust Gates \cite{Liu2016SpatioTemporalLW}} & {\footnotesize \hspace*{0.32cm} 69.20} & {\footnotesize \hspace*{0.24cm} 77.70}   \\
{\footnotesize \cellcolor{gray!50} Two-Layer P-LSTM \cite{Shahroudy2016NTURA}} & {\footnotesize \cellcolor{gray!50}  \hspace*{0.25cm} 62.93} & {\footnotesize \cellcolor{gray!50}  \hspace*{0.17cm} 70.27} \\
{\footnotesize   \hspace*{0.01cm} Dynamic Skeletons \cite{Hu2015JointlyLH}} & {\footnotesize   \hspace*{0.32cm} 60.20} & {\footnotesize   \hspace*{0.24cm} 65.20} \\
{\footnotesize   \cellcolor{gray!50} STA-LSTM \cite{Song2017AnES} } & {\footnotesize \cellcolor{gray!50}  \hspace*{0.25cm} 73.40} & {\footnotesize \cellcolor{gray!50}   \hspace*{0.17cm} 81.20} \\
{\footnotesize   \hspace*{0.01cm} Depth and Skeleton Fusion \cite{Rahmani_2017_ICCV}} & {\footnotesize  \hspace*{0.32cm} 75.20} & {\footnotesize   \hspace*{0.24cm} 83.10} \\
\hline
{\footnotesize  \cellcolor{gray!50} SPMF Inception-ResNet-121} & {\footnotesize  \cellcolor{gray!50}  \hspace*{0.25cm} \textbf{77.02}} & {\footnotesize \cellcolor{gray!50}  \hspace*{0.17cm} 82.13 }  \\
{\footnotesize  \hspace*{0.01cm} SPMF Inception-ResNet-222} & {\footnotesize   \hspace*{0.32cm} \textcolor{blue}{\textbf{78.89}} } & {\footnotesize \hspace*{0.24cm} \textcolor{blue}{\textbf{86.15}}}  \\
{\footnotesize \cellcolor{gray!50} SPMF Inception-ResNet-242} & {\footnotesize \cellcolor{gray!50} \hspace*{0.25cm}  \textbf{77.24}} & {\footnotesize \cellcolor{gray!50} \hspace*{0.17cm} \textbf{83.45} }  \\
\hline
\hline
\end{tabular}
\end{table}
\begin{figure}[htb]
\centering
 \centerline{\includegraphics[width=8.6cm,height=3.1cm]{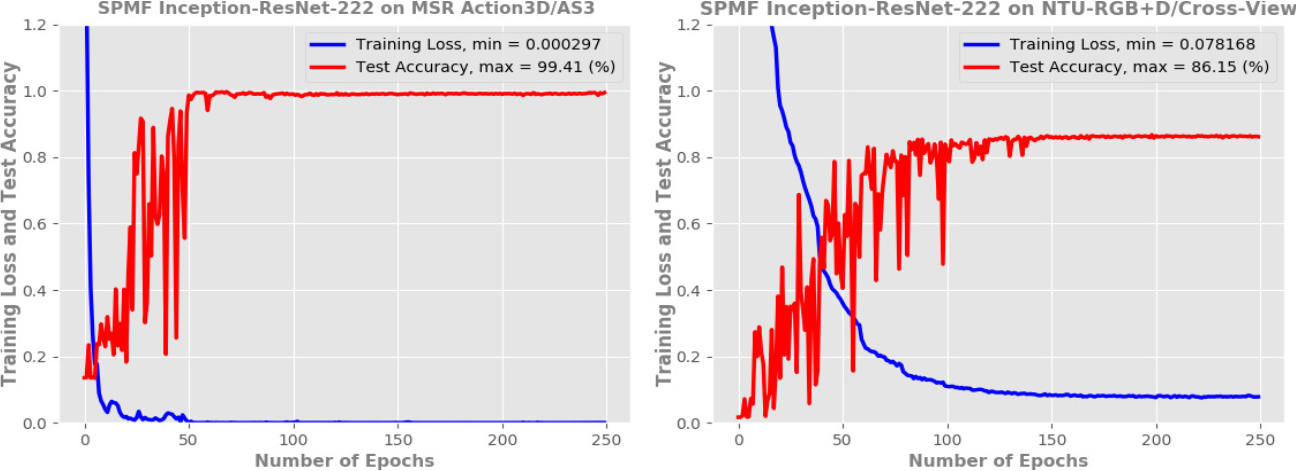}}
\caption{Training loss and test accuracy of SPMF-Inception-ResNet-222 on MSR Action3D and NTU-RGB+D datasets.}
\label{fig:accuracy}
\end{figure}
\begin{figure}[H]
\begin{minipage}[b]{1.0\linewidth}
  \centering
  \centerline{\includegraphics[width=8.6cm,height=1.72cm]{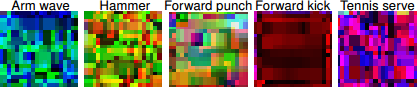}}
\end{minipage}
\caption{Visualizing intermediate features generated by Inception-ResNet-222 after feeding several SPMFs into the network. These SPMFs come from samples in the MSR Action3D dataset \cite{Li2010ActionRB}. Best viewed in color.}
\label{fig:res}
\end{figure}
\subsection{Training and prediction time}
We take the NTU-RGB+D dataset with cross-view settings and SPMF-Inception-ResNet-222 network for illustrating the computational efficiency of our learning framework. With the implementation in Python on a single GeForce GTX Ti GPU, no parallel processing, the training phase takes \textbf{1.85$ \times $10$^{\textbf{-3}}$} second per sequence. While the testing phase, including the time for encoding skeletons into color images and classification, takes \textbf{0.128} second per sequence. This speed is fast enough to the needs of many real-time applications. These results verify the effectiveness of the proposed method, not only in terms of accuracy, but also in terms of computational cost.
\section{Conclusion and Future Work}
\label{sect:5}
In this paper, we introduced a new method for recognizing human actions from skeletal data. A novel skeleton-based representation, namely SPMF, is proposed for encoding spatial-temporal dynamics of skeleton joints into color images. Deep convolutional neural networks based on Inception Residual architecture are then exploited to learn and recognize actions from obtained image-based representations. Experiments on two publicly available benchmark datasets have demonstrated the effectiveness of the proposed representation as well as feature learning networks. We are currently expanding this study by improving the color encoding algorithm through an image enhancement method in order to generate more discriminative features contained in the image representations.
\bibliographystyle{IEEEbib}

\end{document}